\documentclass[letterpaper, conference, 10 pt]{ieeeconf}  
\usepackage[left=54pt,right=54pt,top=54pt,bottom=54pt]{geometry}
\IEEEoverridecommandlockouts                              
\usepackage{graphics} 
\usepackage{epsfig} 
\usepackage{mathptmx} 
\usepackage{times} 
\usepackage{amsmath} 
\usepackage{amssymb}  
\usepackage{booktabs} 
\usepackage{cite}
\usepackage{amsfonts}
\usepackage{amsbsy}
\usepackage{placeins}
\usepackage{multirow}
\usepackage{tabularx}
\usepackage{bm}
\usepackage{url}

\title{\LARGE \bf
Control Pneumatic Soft Bending Actuator with Online Learning Pneumatic Physical Reservoir Computing}

\author{Junyi Shen, Tetsuro Miyazaki, and Kenji Kawashima
\thanks{*This work was supported by Japan Society for the Promotion of Science (JSPS) under grant No. KAKENHI 21H04544 and Bridgestone Corporation.}
\thanks{The authors are with the Department of Information Physics and Computing, The University of Tokyo, 7-3-1 Hongo, Bunkyo-Ku, Tokyo, Japan}
}

\begin{document}

\maketitle
\thispagestyle{empty}
\pagestyle{empty}

\begin{abstract}

The intrinsic nonlinearities of soft robots present significant control but simultaneously provide them with rich computational potential. Reservoir computing (RC) has shown effectiveness in online learning systems for controlling nonlinear systems such as soft actuators. Conventional RC can be extended into physical reservoir computing (PRC) by leveraging the nonlinear dynamics of soft actuators for computation. This paper introduces a PRC-based online learning framework to control the motion of a pneumatic soft bending actuator, utilizing another pneumatic soft actuator as the PRC model. Unlike conventional designs requiring two RC models, the proposed control system employs a more compact architecture with a single RC model. Additionally, the framework enables zero-shot online learning, addressing limitations of previous PRC-based control systems reliant on offline training. Simulations and experiments validated the performance of the proposed system. Experimental results indicate that the PRC model achieved superior control performance compared to a linear model, reducing the root-mean-square error (RMSE) by an average of over $37\%$ in bending motion control tasks. The proposed PRC-based online learning control framework provides a novel approach for harnessing physical systems' inherent nonlinearities to enhance the control of soft actuators.

\end{abstract}

\section{INTRODUCTION}

Pneumatic artificial muscles (PAMs), with the McKibben design \cite{chou1996measurement}, are widely used soft actuators because of their high force-to-weight ratio. While PAMs typically perform linear motions, modifications like introducing physical constraints into the standard structure of PAMs \cite{ quevedo2023design, shen2023trajectory} have enabled PAM-based actuators to perform bending movements.

As a result of the soft materials' nonlinear properties, soft actuators, e.g., PAMs, present control challenges \cite{nakajima2015information}. One prominent characteristic of PAMs and other soft actuators is hysteresis, which is a time-dependent nonlinear dynamics. The hysteresis of PAMs results from the Coulomb friction between mechanical components and the viscoelasticity of soft materials \cite{chou1996measurement}. Researchers have been actively developing methods to precisely control the motions of PAMs \cite{xie2018hysteresis} and other soft actuators \cite{cheng2016adaptive, zou2023generalized} in the presence of hysteresis. While model-based controllers have been widely favored for their analyzability and effectiveness in handling nonlinear systems \cite{cheng2016adaptive, zou2023generalized}, accurately modeling the complex nonlinear dynamics is challenging \cite{waegeman2012feedback}. Meanwhile, proportional-integral-differential (PID) controllers, which have been widely used in model-free control systems for their ease of implementation, often exhibit suboptimal performance when applied to nonlinear plants \cite{liang2023online}.

With the development of machine learning, recurrent neural networks (RNNs), which have powerful capability to capture complex time-dependent nonlinearities, have gained popularity in the modeling \cite{sun2022physics} and control \cite{jiang2022intelligent} of soft actuators. However, RNNs' gradient-based training procedure can be computationally expensive, with a large amount of data and considerable time required. Such a training process also easily meets vanishing and exploding gradient problems.

In light of conventional RNNs' computational complexity, Reservoir Computing (RC) \cite{jaeger2002adaptive}, a variant of RNNs, has been introduced to model time-dependent nonlinear dynamics. In RC models, such as the commonly used Echo State Network (ESN), the input signal is projected into a high-dimensional reservoir state via an input layer. The state is updated by a large and randomly initialized network matrix, termed the reservoir, which typically contains tens to thousands of nodes \cite{moon2019temporal, park2016online}. Notably, the input layer and the reservoir matrix remain fixed after their random initialization, with only the output layer, which processes the reservoir state via linear combination, being adjusted by linear regression methods.

Because of their gradient-free training procedures, RC models are well-suited for real-time applications like RNN-based online learning control systems \cite{liang2023online}. Previous research has exhibited the effectiveness of ESNs combined with online learning algorithms, such as the recursive least square (RLS) \cite{waegeman2012feedback, park2016online, jordanou2019online} or the filtered-X algorithm \cite{liang2023online}, in controlling complex nonlinear systems. Using the ESN-based online learning model to control PAMs has also been explored \cite{xing2012modeling}.

Existing ESN-based online learning control systems typically involve two identical ESN models—one for online adjustment and the other for controlling the plant \cite{waegeman2012feedback}. Considering that ESNs depend on a large reservoir for state updates, these online learning methods may confront real-time application challenges in edge devices with limited computational capacities, as such devices may struggle to handle repeated operations engaging the large reservoir matrices \cite{moon2019temporal, yamazaki2024photonic}.

Given that the reservoir matrix in ESN stays fixed during training and execution, the reservoir part can be replaced by real-world physical systems with nonlinearities and memory capacities, such as PAMs \cite{hayashi2022, akashi2024embedding} and other soft bodies \cite{eder2018morphological, nakajima2015information}. Such substitutabilities have led to the development of Physical Reservoir Computing (PRC). Prior work has shown the use of PRC as a feedforward hysteresis compensation model for controlling the movements of a soft actuator \cite{shen2024control}. As PRCs rely on no large reservoir matrix, they can show significantly reduced computational requirements compared to ESNs \cite{akashi2024embedding, shen2024control}, highlighting their suitability in real-time control applications. However, the PRC-based control system proposed in \cite{shen2024control} relies on offline training with no online adjustment mechanism considered, leaving the utilization of online learning PRC to control soft actuators to be explored.

This paper introduces a new RC-based online learning control system design, which utilizes only one RC model, thus offering a more compact and computationally lightweight structure than existing designs featuring two RC models. Furthermore, the RC model in the proposed system can be either a conventional ESN or a PRC, facilitating the implementation of PRCs in online learning control systems. The proposed system eliminates the training requirement of the existing PRC-based control model \cite{shen2024control}. Its effectiveness is verified by simulations in controlling the output of a simulated nonlinear system and by experiments in controlling the movements of a soft actuator. The simulation and experimental results show that the online learning control system incorporating an RC model outperforms the linear control system in managing the outputs of the simulated nonlinear plant and the soft actuator.

The following content is organized as follows: Section 2 introduces the pneumatic soft actuator used as the control object in this study and presents the ESN and PRC models. Section 3 details the proposed online learning control system architecture. Section 4 and Section 5 provide the simulation and experimental results, respectively. Section 6 concludes.

\section{Soft Actuator and RC Models}

This section introduces the ESN and the PRC models \cite{shen2024control}; the PRC model is based on the dual-PAM pneumatic soft bending actuator detailed in \cite{shen2023trajectory}. Before discussing the two different RC models, we present the pneumatic soft bending actuator, whose motion is the control object of this work.

\subsection{Pneumatic Soft Bending Actuator}

\begin{figure}[tb]
    \centering
\includegraphics[width=0.8\linewidth]{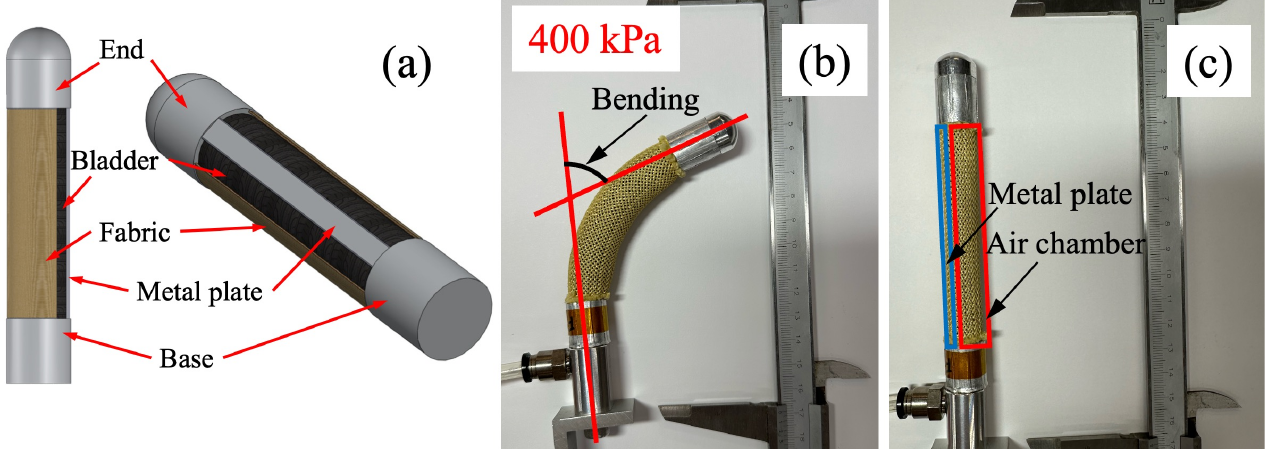}
    \caption{Pneumatic soft bending actuator: (a) Structure; (b) Bending under pressurization; (c) Returning to initial straight shape upon depressurization.}
    \label{fig: pam}
    \vspace{-0.3cm}
\end{figure}

Fig.~\ref{fig: pam}(a) is a schematic of the pneumatic soft actuator's structure, which contains a metallic base, a rubber bladder, a metal end, a metal plate, and an inextensible fabric shield. Following the mechanism of PAMs, the rubber bladder expands radially and contracts axially upon pressurization. The metal plate impedes the actuator's uniformly axial contraction, hence driving its bending movements, as Fig.~\ref{fig: pam}(b) shows. Experiments in \cite{shen2024control} illustrate that the pneumatic actuator's bending output is roughly 60° at the pressurization level of 400 kPa. Upon depressurization, the soft actuator returns to its original straight shape, as Fig.~\ref{fig: pam}(c) exhibits.

The control objective of this work is to manipulate the pneumatic soft bending actuator's actuation pressure such that the actuator's bending angle output $y_k$ closely tracks the reference signal $y^d_k$ at each discrete time step $k$. Considering the sensor noise in practical closed-loop control systems, the feedback reading of the actual output $y$ is denoted as $\tilde{y}$.

The nonlinearities and hysteresis of the soft actuator shown in Fig.~\ref{fig: pam} have been detailed in \cite{shen2024control}. As this study employs an online learning control system, where neither the experimental data for offline model training nor prior knowledge about the controlled actuator's dynamics is required, an exhibition of the soft actuator's nonlinearities is omitted here for brevity.

\subsection{Echo State Network (ESN) Model}

\begin{figure}[tb]
    \centering
\includegraphics[width=0.8\linewidth]{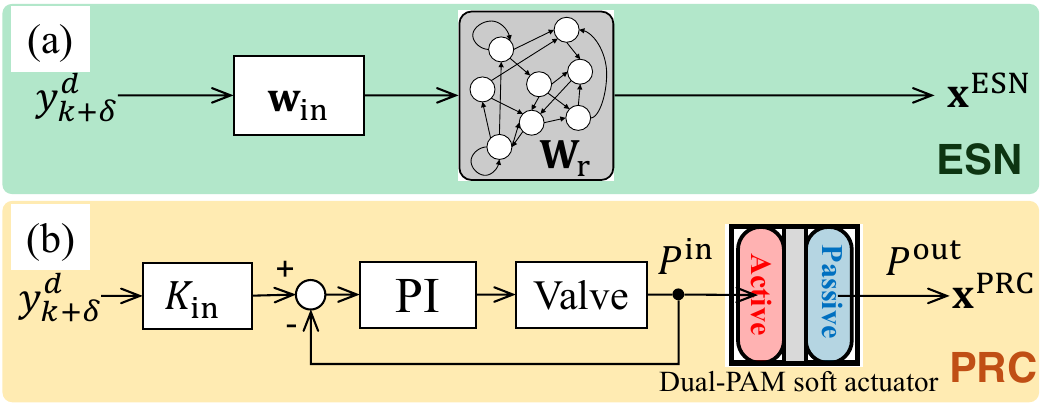}
    \caption{Schematic of RC models: (a) The ESN model; (b) The PRC model, the red and blue boxes indicate the active and passive PAM shown in Fig.~\ref{fig: reservoir}.}
    \label{fig: esn_prc}
    \vspace{-0.3cm}
\end{figure}

Training neural networks (NNs) with direct input-output data reversal may not yield the optimal results \cite{zhou2020deep}. Previous works have employed future reference signals \cite{waegeman2012feedback, park2016online, jordanou2019online}, which are often available in practical applications \cite{zhou2020deep}, as inputs to the ESN model in online learning control systems. Motivated by the success of previous works, we design the input to the ESN model in our control system as $y^d_{k+\delta}$, the $\delta$-step future reference signal. The schematic of the ESN model is shown in Fig.~\ref{fig: esn_prc}(a), where the input $y^d_{k+\delta}$ is projected into the ESN's high-dimensional reservoir state $\mathbf{x}^\text{ESN} \in \mathbb{R}^{N}$, as:
\begin{equation}
    \mathbf{x}^\text{ESN}_{k+1} = (1-\gamma)\mathbf{x}^\text{ESN}_k + \gamma \tanh(\mathbf{W}^\text{r} \mathbf{x}^\text{ESN}_k + \mathbf{w}^\text{in}y^d_{k+\delta}),
\label{eq: esn update}
\end{equation}
where $\gamma \in (0,1]$ is the leaky rate, $\mathbf{W}^\text{r} \in \mathbb{R}^{N\times N}$ is the reservoir matrix of size $N$, $\mathbf{w}^\text{in} \in \mathbb{R}^{N}$ is the input layer, and $\tanh(\cdot)$ is the hyperbolic tangent activation function. Both $\mathbf{W}^\text{r}$ and $\mathbf{w}^\text{in}$ are randomly initialized, scaled according to respectively selected coefficients, and fixed during online learning. The high-dimensional state $\mathbf{x}^\text{ESN}$ is subsequently combined with linear weights contained in the output layer (introduced in the following Section 3) to generate the ESN model’s output.

\subsection{Physical Reservoir Computing (PRC) Model}

\begin{figure}[tb]
    \centering
\includegraphics[width=0.8\linewidth]{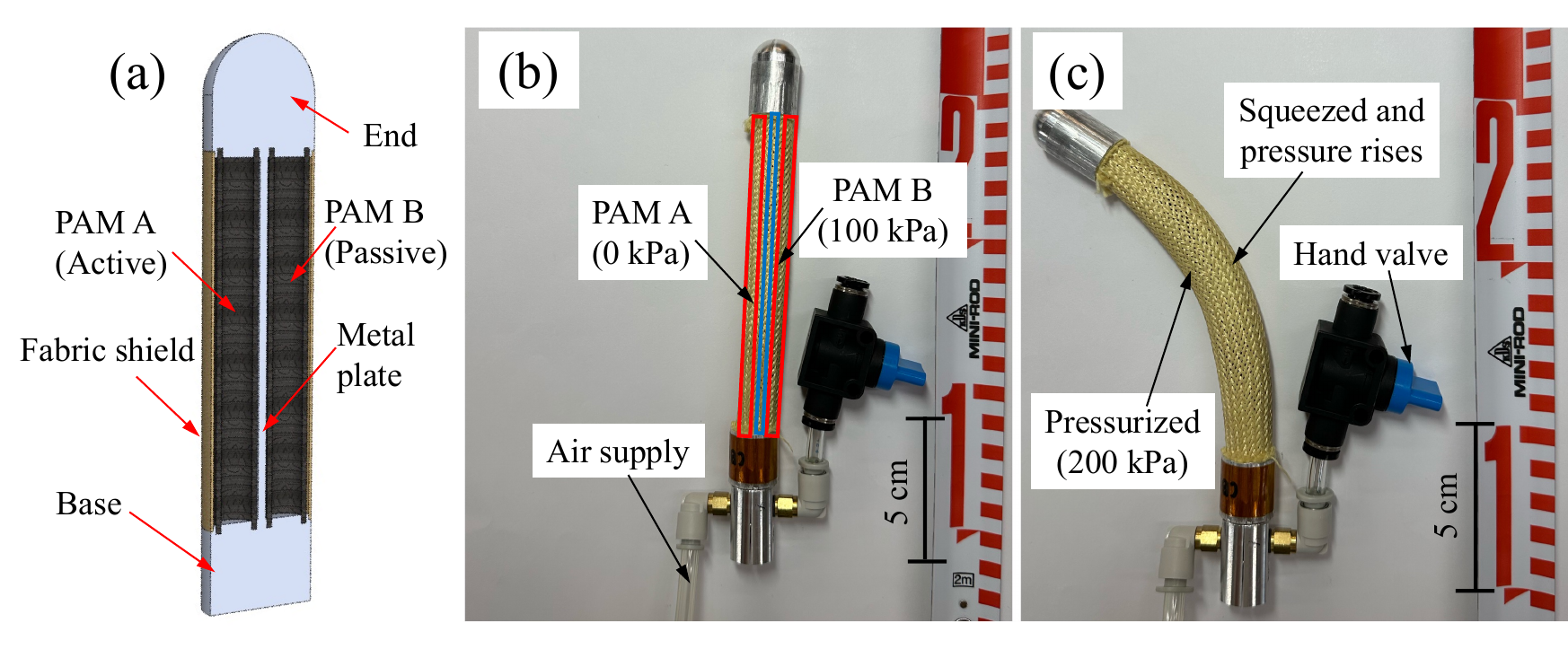}
    \caption{Dual-PAM actuator as a physical reservoir: (a) Structural design; (b) Active PAM (PAM A) pressurized and passive PAM (PAM B) pre-pressurized and sealed; (c) Actuation of PAM A drives the fabric to squeeze the pre-pressurized and sealed PAM B, increasing its internal pressure.}
    \label{fig: reservoir}
    \vspace{-0.3cm}
\end{figure}

The dual-PAM pneumatic actuator, serving as the physical reservoir in this work, is shown in Fig.~\ref{fig: reservoir}. Its structure, shown by Fig.~\ref{fig: reservoir}(a), closely mirrors the design in Fig.~\ref{fig: pam}(a), with the addition of a second rubber bladder and the central metal plate flanked by the two antagonistically located air chambers. In this setup, one air chamber, the PAM A in Fig.~\ref{fig: reservoir}, functions as an active PAM, which is actuated by the customized pressure input. The other air chamber, the PAM B in Fig.~\ref{fig: reservoir}, acts as a passive PAM, which is initially pressurized to 100 kPa and closely sealed by a hand valve, as illustrated in Fig.~\ref{fig: reservoir}(b). The dual-PAM actuator's functionality as a physical reservoir has been thoroughly discussed in \cite{shen2024control} and is omitted here for brevity. The schematic of the PRC model utilizing this dual-PAM actuator is shown in Fig.~\ref{fig: esn_prc}(b). Similar to the ESN model, the input to the PRC is $y^d_{k+\delta}$, which is transformed by a conversion factor $K^\text{in}$ [kPa/deg] into the actuation pressure of the active PAM, as follows:
\begin{equation}
    P^\text{in}_k = K^\text{in} y^d_{k+\delta},
\label{eq: convert}
\end{equation}
where $P^\text{in}$ [kPa] is the actuation pressure of the active PAM, which is regulated by a proportional valve and a proportional-integral (PI) controller. The pressurization and inflation of the active PAM cause the inextensible fabric shield to compress the pre-pressurized and sealed passive PAM, thereby altering its internal pressure $P^\text{out}$ [kPa], as shown in Fig.~\ref{fig: reservoir}(c). The dynamics of this transformation are defined in this work as:
\begin{equation}
    P^\text{out}_{k+1} = \mathcal{F}_r\left(P^\text{out}_k, \ P^\text{in}_k \right) = \mathcal{F}_r\left(P^\text{out}_k, \ K^\text{in} y^d_{k+\delta} \right),
\label{eq: nonlinear_mapping}
\end{equation}
where $\mathcal{F}_r(\cdot)$ denotes the nonlinear mapping with memory capacity from $P^\text{in}$ to $P^\text{out}$ \cite{shen2024control}. The valve's dynamics and any errors in the regulation of pressure $P^\text{in}$ are included in the overall dynamics described by $\mathcal{F}_r(\cdot)$. Note that the exact mathematical description of $\mathcal{F}_r(\cdot)$ is difficult to obtain and not required in the PRC's use. The passive PAM's internal pressure $P^\text{out}$ serves as the readout of the physical reservoir's state. The ESN's leaky rate $\gamma$ is emulated by a low-pass filter with a filter factor $\epsilon \in (0,1]$ in the PRC model, as follows:
\begin{equation}
\Breve{P}^\text{out}_k = \epsilon P^\text{out}_k + (1-\epsilon) \Breve{P}^\text{out}_{k-1},
\end{equation}
where $\Breve{P}^\text{out}$ denotes the filtered state readout of the physical reservoir. Different from the ESN, whose state $\mathbf{x}^\text{ESN}$ is fully accessible and can be defined with arbitrary dimensionality, while the physical reservoir has a high-dimensional real-world structure, the readout of its state is represented by a scalar $\Breve{P}^\text{out}$. Due to the absence of explicitly tunable dimensionality, the computational performance of the physical reservoir is much less flexible than that of ESNs \cite{lee2024task}. Previous works have handled this limitation by using temporal sequences of the physical reservoir's outputs (state readouts) with arbitrary tap sizes as compensation \cite{akashi2024embedding, hayashi2022}. Similarly, in this work, we collect the filtered physical reservoir's states into temporal sequences of tap size $n_u$, represented as $\mathbf{x}^\text{PRC}_k = [\Breve{P}^\text{out}_k, \Breve{P}^\text{out}_{k-1}, \cdots, \Breve{P}^\text{out}_{k-n_u+1}]^\top \in \mathbb{R}^{n_u}$, where $n_u$ can be arbitrarily defined. While increasing the dimensionality of the reservoir state, e.g., by including additional observables such as the physical reservoir's shape or bending angle, could potentially enhance the PRC's computational power, the environmental robustness of the inner pressure $\Breve{P}$, as verified in \cite{shen2024control}, serves as the motivation for its selection. The state $\mathbf{x}^\text{PRC}$ is then combined with the linear weights contained in the output layer (discussed in the following Section 3) to generate the PRC model's output. Notably, the physical reservoir's memory capacity is primarily governed by the soft actuator's hysteresis dynamics \cite{caremel2024hysteretic}, which dominate the reservoir's time-dependent properties unless time-window length $n_u$ is selected to match the time scale of the hysteresis dynamics.

\section{Online Learning Control System}

\begin{figure}[tb]
    \centering
\includegraphics[width=0.8\linewidth]{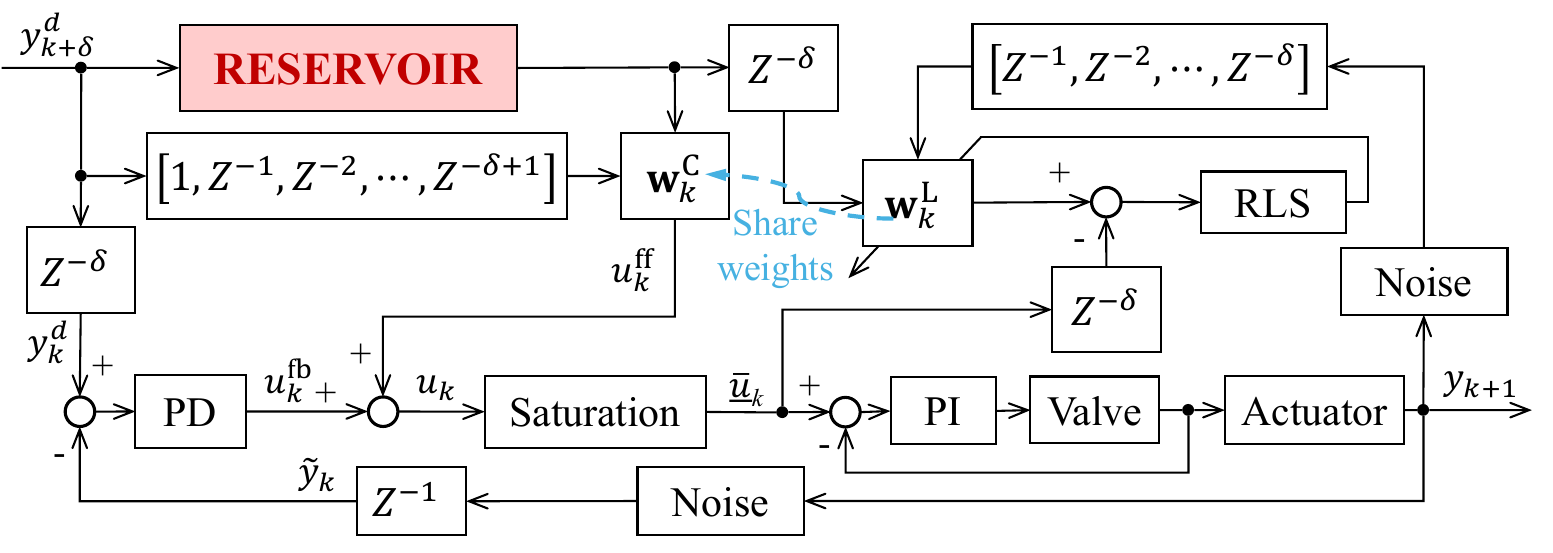}
    \caption{Online learning control system. $Z$ symbol denotes the Z-transform.}
    \label{fig: system}
    \vspace{-0.3cm}
\end{figure}

Fig.~\ref{fig: system} illustrates the online learning control system used to regulate the pneumatic soft actuator's bending motion. The control system adopts a two-degree-of-freedom design, with an RC model as the feedforward component and a proportional-differential (PD) controller for feedback adjustment. The primary control workflow aligns with that in \cite{shen2024control}, with the vital distinction being that, in this work, the RC model is trained online, whereas the model in \cite{shen2024control} is trained offline and fixed during execution. Moreover, in contrast to previous online learning control systems using two identical RC models, such as \cite{waegeman2012feedback}, the proposed system features a more compact architecture with only one single RC model used.

As RCs are typically trained by linear regression methods like least squares or ridge regression, the RLS algorithm \cite{waegeman2012feedback} is utilized for the proposed control system's online learning. In the proposed control system design, the $\delta$-step future reference signal $y_{k+\delta}^\text{d}$ is input into the reservoir, which can be either an ESN or a PRC, to update the reservoir state $\mathbf{x} \in \mathbb{R}^r$, where $r$ denotes the state vector's size. The state $\mathbf{x}$ denotes either $\mathbf{x}^\text{ESN}$ when the reservoir is an ESN or $\mathbf{x}^\text{PRC}$ when it is a PRC. Consequently, the vector size $r$ is $N$ for $\mathbf{x}^\text{ESN}$ and $n_u$ for $\mathbf{x}^\text{PRC}$. The state $\mathbf{x}$ is then combined with the reference signal's future information $\mathbf{y}^\text{d}_k = [y_{k+1}^\text{d}, y_{k+2}^\text{d}, \cdots, y_{k+\delta}^\text{d}]^\top \in \mathbb{R}^\delta$ to form an extended state vector as $[\mathbf{x}^\top, {\mathbf{y}^\text{d}}^\top]^\top \in \mathbb{R}^{r+\delta}$. The extended state is processed by the output layer $\mathbf{w}^\text{C} \in \mathbb{R}^{r+\delta}$ to produce the feedforward control output $u^\text{ff}_k$ [kPa] following:
\begin{equation}
    u^\text{ff}_k = \mathbf{w}^\text{C}_k [\mathbf{x}_k^\top, \ {\mathbf{y}_k^\text{d}}^\top]^\top.
\label{eq: ff output}
\end{equation}
An additional output layer $\mathbf{w}^\text{L}\in \mathbb{R}^{r+\delta}$, which is initialized identically to $\mathbf{w}^\text{C}$, is adjusted using the RLS algorithm with the historical input-output data pairs at each time step $k$ \cite{waegeman2012feedback}:
\begin{align}
    e^r_{k} &= \mathbf{w}^\text{L}_{k-1}[\mathbf{x}_{k-\delta}^\top, \ \tilde{\mathbf{y}}_{k-\delta}^\top]^\top - \bar{\underline{u}}_{k-\delta} \\
    \mathbf{w}^\text{L}_{k} &= \mathbf{w}^\text{L}_{k-1}-e^r_{k}\left(\mathbf{P}_{k} \ [\mathbf{x}_{k-\delta}^\top, \ \tilde{\mathbf{y}}_{k-\delta}^\top] \right)^\top \\
    \mathbf{P}_{0} &= \frac{\mathbf{I}}{\alpha}  \\
    \mathbf{P}_{k} &= \frac{\mathbf{P}_{k-1}}{\lambda} - \frac{\mathbf{P}_{k-1}[\mathbf{x}_{k-\delta}^\top, \ \tilde{\mathbf{y}}_{k-\delta}^\top]^\top \ [\mathbf{x}_{k-\delta}^\top, \ \tilde{\mathbf{y}}_{k-\delta}^\top] \ \mathbf{P}_{k-1}}{\lambda \left( \lambda + [\mathbf{x}_{k-\delta}^\top, \ \tilde{\mathbf{y}}_{k-\delta}^\top] \mathbf{P}_{k-1} [\mathbf{x}_{k-\delta}^\top, \ \tilde{\mathbf{y}}_{k-\delta}^\top]^\top \right)},
\end{align}
where $\bar{\underline{u}}$ denotes the actual input (with saturation) to the controlled soft actuator, $\tilde{\mathbf{y}}_{k-\delta} = [\tilde{y}_{k-\delta+1}, \tilde{y}_{k-\delta+2}, \cdots, \tilde{y}_k]^\top \in \mathbb{R}^\delta$ is the temporal sequence of the feedback signal $\tilde{y}$, $0 < \alpha$ is the learning rate, $0 \ll \lambda < 1$ is the forgetting factor, $\mathbf{I} \in \mathbb{R}^{r \times r}$ is the identity matrix, and $\mathbf{P}_{k}$ is the running estimate of the Moore-Penrose pseudoinverse matrix $\left([\mathbf{x}_{k-\delta}^\top, \ \tilde{\mathbf{y}}_{k-\delta}^\top] [\mathbf{x}_{k-\delta}^\top, \ \tilde{\mathbf{y}}_{k-\delta}^\top]^\top + \alpha \mathbf{I} \right)^{-1}$ \cite{waegeman2012feedback}. To construct the online learning RC control system, the output layer $\mathbf{w}^\text{C}$ is initialized as $\mathbf{0} \in \mathbb{R}^{r+\delta}$, then copied to $\mathbf{w}^\text{L}$, which is adjusted by the RLS algorithm at each time step, and then transferred to $\mathbf{w}^\text{C}$ to produce the feedforward output as described by (\ref{eq: ff output}).

The feedforward RC model's output $u^\text{ff}$ is then combined with the PD controller's feedback adjustment $u^\text{fb}$ [kPa] to produce the overall controller output $u$ [kPa] as follows:
\begin{equation}
    u_{k} = \bar{\underline{u}}_{k-1} + u^\text{ff}_{k} - u^\text{ff}_{k-1} + u^\text{fb}_{k},
\label{eq: overall output}
\end{equation}
where the feedback adjustment $u^\text{fb}$ is computed based on the feedback control error defined by $\tilde{e} = y^d-\tilde{y}$ [deg], as follows:
\begin{equation}
    u^\text{fb}_{k} = K^P \ \tilde{e}_k + K^D  \ \frac{\tilde{e}_k-\tilde{e}_{k-1}}{\tau},
\label{eq: fb control}
\end{equation}
where $\tau$ [s] is the sampling interval, $0 \leq K^P$ [kPa/deg] is the proportional (P) gain, and $0 \leq K^D$ [kPa$\cdot$s/deg] is the differential (D) gain. The P and D gain values can be roughly tuned, as the feedback controller's contribution to the online learning control system primarily lies in guiding the RLS algorithm to adjust the RC model with more suitable learning data rather than directly controlling the actuator \cite{shen2024improving}. The overall output $u$ is then passed through a saturation function $f_\text{sat}(\cdot)$ to produce the final input $\bar{\underline{u}}$ [kPa] to the actuator as:
\begin{equation}
    \bar{\underline{u}}_k = f_\text{sat}(u_k).
\end{equation}
The input $\bar{\underline{u}}$ is regulated by a PI controller and a proportional valve, as shown in Fig.~\ref{fig: system}. The PI controller is assumed to perform ideally, as pressure control is not the primary focus of this study and typically achieves satisfactory performance compared to the more challenging task of controlling the soft actuator's motion. It should be noted that the differential operation in (\ref{eq: fb control}) is adopted in this work for simplicity. It is often not recommended in practice due to its tendency to amplify noises, which may destroy the system's stability.

Compared with the method adopted in \cite{shen2024control}, the proposed system eliminates the prior training of the RC model, thereby reducing the effort required for data acquisition and model learning. Furthermore, unlike online learning systems with two RC models and perform two state updates at each time step \cite{waegeman2012feedback, park2016online, jordanou2019online, shen2024improving}, the proposed system employs one single reservoir, with only one state update operation conducted at each time step, making it more computationally lightweight. Additionally, existing designs using two identical RC models present challenges in replacing conventional RC models with PRCs, as achieving the identicalness between two physical systems is practically impossible \cite{moon2019temporal}. In contrast, the proposed system, which uses only a single RC model, allows seamless replacement with physical reservoirs.

\section{Simulations}

\subsection{Simulation Setup}

We used simulations to evaluate the performance of our control system design shown in Fig.~\ref{fig: system}. Given the complexities associated with modeling the soft actuator shown in Fig.~\ref{fig: pam} and the pneumatic physical reservoir in Fig.~\ref{fig: reservoir}, we employed an ESN model as the reservoir part in the control system and used such a control system to regulate the output of a simulated nonlinear system following various reference signals. For simplicity, all variables are considered unitless in the simulations. The design of the simulated nonlinear system follows that used in \cite{park2016online}. The system's dynamics is:
\begin{equation}
y_{k+1} = \frac{y_k}{1+y^2_k} + \bar{\underline{u}}^3_k,
\end{equation}
where $\bar{\underline{u}}$ and $y$ is the input and output of the simulated system. Despite the identity in system dynamics, the online learning ESN model in \cite{park2016online} was pre-trained, while the ESN model in our simulations was not trained prior to its application.

The hyperparameter settings were determined through trial and error. The ESN model's reservoir size $N$ was 50, and the leaky rate $\gamma$ was 0.8. The reservoir $\mathbf{W}^\text{r}$ was initialized with $\mathcal{U}(-0.5, 0.5)$ elements, where $\mathcal{U}(a, b)$ represents a random value drawn from a uniform distribution between $a$ and $b$, $a < b$. After initialization, $\mathbf{W}^\text{r}$ was scaled such that its largest absolute eigenvalue (also known as the spectral radius) was 0.8. The input layer $\mathbf{w}^\text{in}$ was randomly initialized with $\mathcal{U}(-1, 1)$ elements, then scaled by a factor of 1 (i.e., no scaling). The tap size $\delta$ was 5, and the ESN's state $\mathbf{x}^\text{ESN}$ was initialized by running (\ref{eq: esn update}) with zero input for 100 steps, which is known as the washout process. The RLS's learning rate $\alpha$ was set to 1, and the forgetting factor $\lambda$ was $1-1\times10^{-6}$. The sampling interval $\tau$ was set to $5\times10^{-3}$. The PD controller’s proportional gain $K^P$ was $1\times10^{-4}$, and the differential gain $K^D$ was $1\times10^{-6}$. The low values for the P and D gains were chosen to prevent the feedback controller from dominating the control of the nonlinear system's output, thus more clearly illustrating the contribution of the feedforward RC model. Despite the low gains, including the feedback controller in the system is essential to accelerate the convergence of the online learning feedforward model \cite{shen2024improving}, particularly given that the ESN was not pre-trained in our simulations. The saturation function $f_\text{sat}(\cdot)$ in our simulations was defined as no saturation, i.e.:
\begin{equation}
    \bar{\underline{u}}_k = f_\text{sat}(u_{k}) = u_{k}.
\label{eq: simu sat}
\end{equation}
The feedback sensor noise in simulations was simulated by:
\begin{equation}
    \tilde{y}_k = y_k + \mathcal{N}(0, 0.1),
\end{equation}
where $\mathcal{N}(a, b)$ denotes a random value drawn from a normal distribution with mean value $a$ and standard deviation $b$. The noisy signal $\tilde{y}$ was used for feedback control in simulations and the actual output $y$ was used to illustrate the results.

For performance comparison, we evaluated two additional control system designs: an online learning control system without the ESN model (referred to as the Linear+PD method) and a standalone PD controller without the feedforward component $u^\text{ff}$ (referred to as the PD method). The proposed control system in Fig.~\ref{fig: system}, which combines the ESN feedforward model with the PD controller, is termed the ESN+PD method in the following discussion. The Linear+PD method was implemented by replacing the extended state $[\mathbf{x}_k^\top, \ {\mathbf{y}_k^\text{d}}^\top]^\top$ and $[\mathbf{x}_{k-\delta}^\top, \ \tilde{\mathbf{y}}_{k-\delta}^\top]^\top$ used in the ESN+PD with $\mathbf{y}_k^\text{d}$ and $\tilde{\mathbf{y}}_{k-\delta}$ for feedforward control and online adjustment, respectively. The hyperparameters of the two comparative control methods were identical to the setup described above. Each simulation was independently repeated 100 times with the same group of random seeds to account for the randomness. In presenting the results, solid lines show the average of the 100 runs, and shaded areas denote the standard deviations. All simulations were executed on an Apple M1 CPU with Python 3.10.0, Numpy 1.23.5, and Scipy 1.10.1. The simulation code used in this study is available at \url{https://github.com/JonyeeShen/Online-Learning-RC-Control-RoboSoft2025}.

\subsection{Simulation Results}

\begin{figure}[tb]
    \centering
\includegraphics[width=0.9\linewidth]{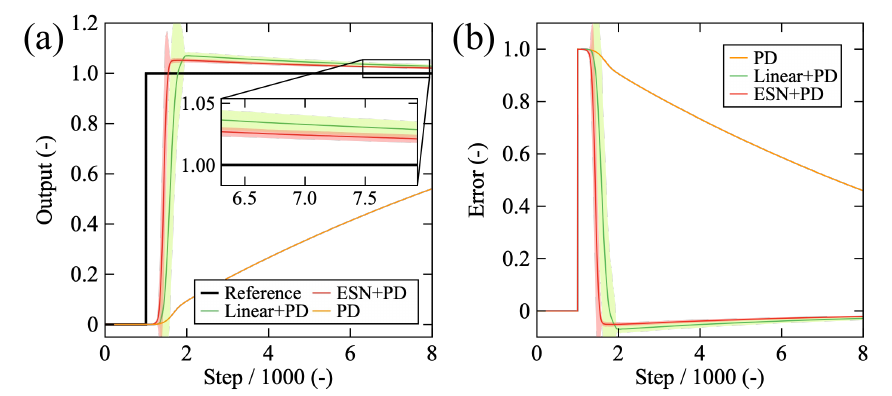}
    \caption{Simulation results of tracking a step signal: (a) Tracking results of different control methods; (b) Control errors of different methods.}
    \label{fig: sim_step}
    \vspace{-0.3cm}
\end{figure}

We evaluated the step responses of the different control methods, and the results are presented in Fig.\ref{fig: sim_step}. As shown in the results, the ESN+PD method exhibited faster rising, reduced overshoot, and quicker convergence with smaller final state control errors compared to the Linear+PD method. On the other hand, due to the slow response caused by the intentionally low feedback gains and the absence of the feedforward element, the PD method was unable to closely align the nonlinear system’s output with the reference signal.

\begin{figure}[tb]
    \centering
\includegraphics[width=0.9\linewidth]{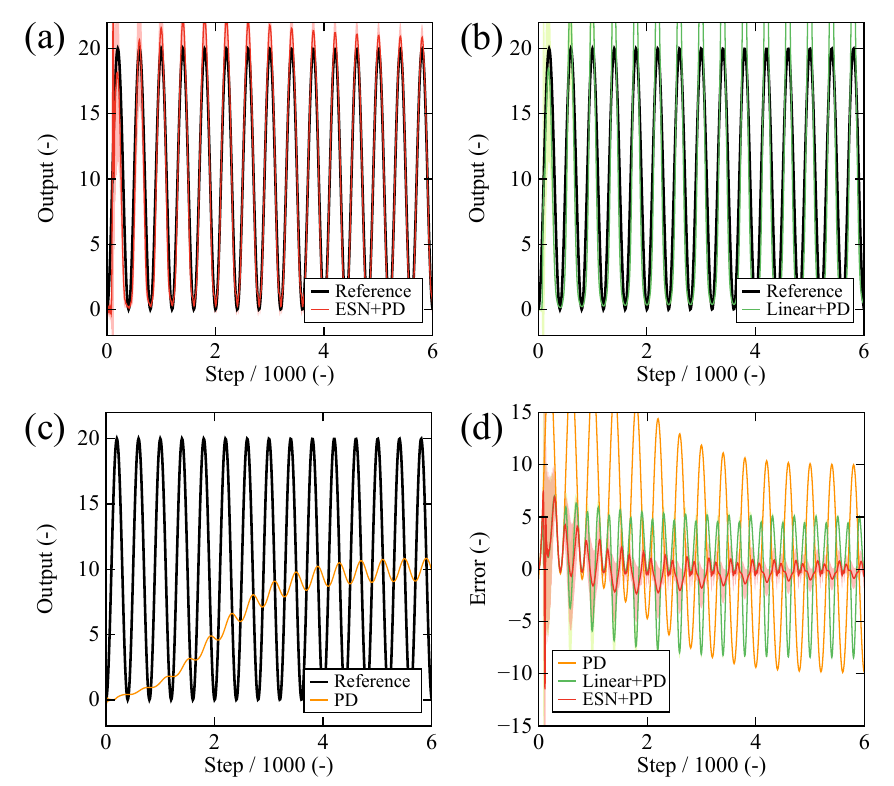}
    \caption{Simulation results for tracking a sine waveform: (a) ESN+PD; (b) Linear+PD; (c) PD; (d) Control errors of different methods.}
    \label{fig: sim_sine}
    \vspace{-0.3cm}
\end{figure}

We evaluated the performance of different control methods in regulating the nonlinear system's output to follow a periodic sine signal, as shown in Fig.~\ref{fig: sim_sine}. Both the Linear+PD and ESN+PD methods successfully tracked the varying reference signal, with the ESN+PD method demonstrating superior performance, characterized by reduced control errors compared to the Linear+PD method. As the feedforward model was entirely untrained, significant control errors were observed during the initial stage (approximately the first 1,000 steps) from the results of both the ESN+PD and the Linear+PD methods, which then gradually decreased to a limited range. Due to its slow response, the PD method failed to control the nonlinear system's output following this reference signal.

\begin{figure}[tb]
    \centering
\includegraphics[width=0.9\linewidth]{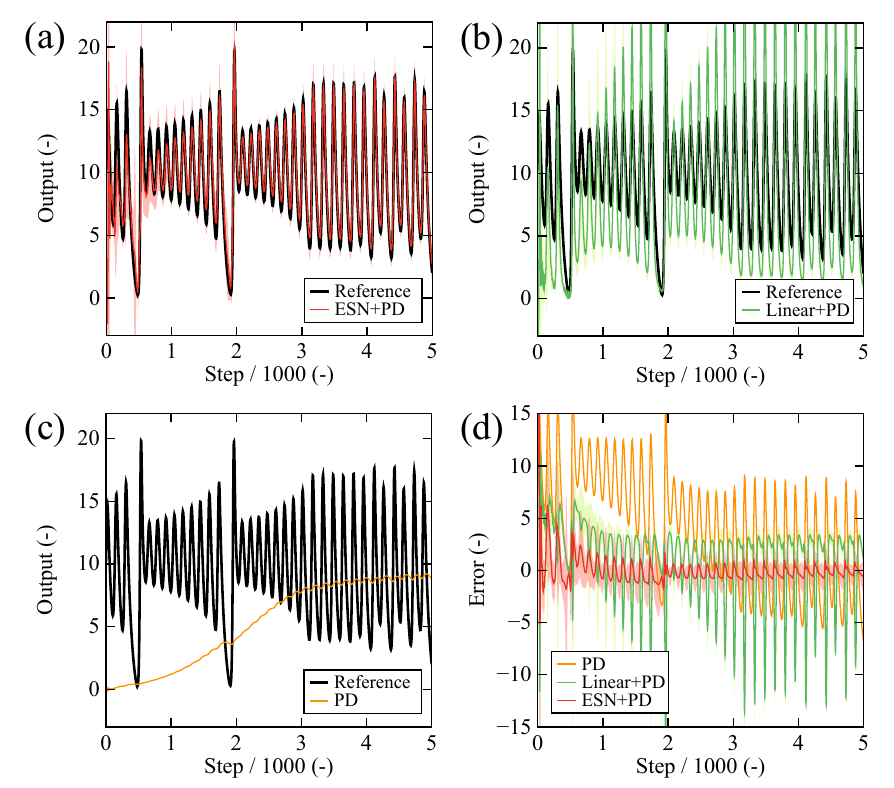}
    \caption{Simulation results for tracking a complex waveform: (a) ESN+PD; (b) Linear+PD; (c) PD; (d) Control errors of different methods.}
    \label{fig: sim_lorenz}
    \vspace{-0.3cm}
\end{figure}

We also evaluated the three control methods in tracking a complex reference signal, with the results shown in Fig.~\ref{fig: sim_lorenz}. Similar to the previous tracking test, both the Linear+PD and ESN+PD methods were able to follow the reference signal's frequent and complex variation. The ESN+PD method exhibited faster convergence and smaller control errors compared to the Linear+PD method, whose unsatisfactory performance can be attributed to the linear feedforward model's disabilities in managing the nonlinear system's output. Similar to the results of previous simulations, the standalone PD controller failed to control the nonlinear system's output following the reference signal in the absence of the feedforward component, highlighting the feedforward (either Linear or ESN) element's crucial contribution to the overall control system.

\begin{figure}[tb]
    \centering
\includegraphics[width=0.9\linewidth]{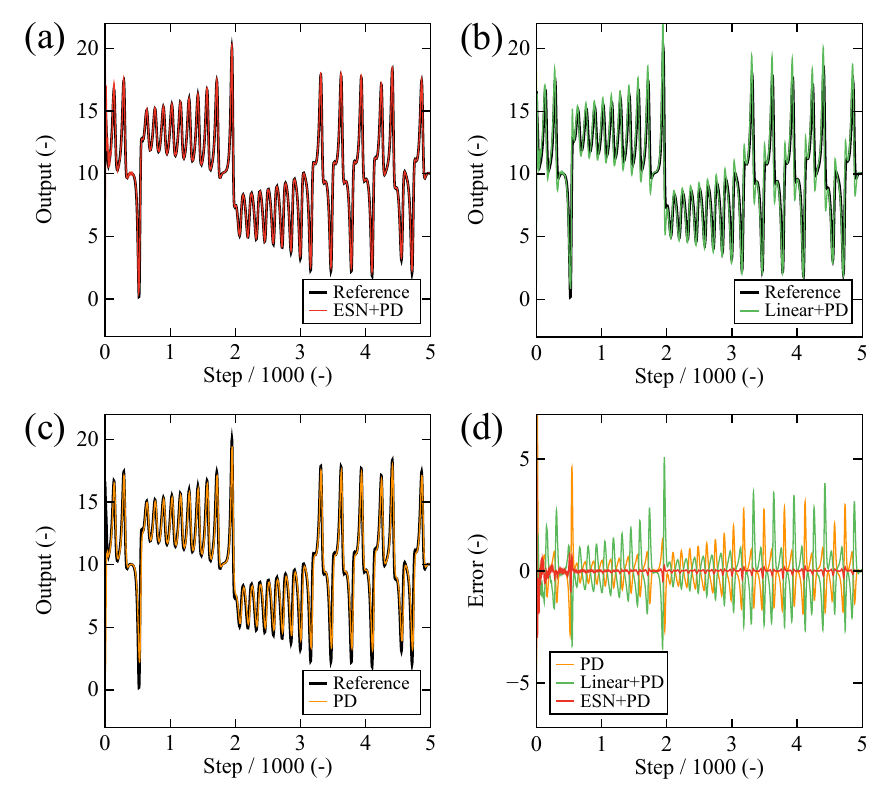}
    \caption{Simulation results for tracking a complex signal with better-selected PD feedback gains: (a) ESN+PD; (b) Linear+PD; (c) PD; (d) Control errors. The Linear and the ESN models were not pre-trained before the simulation.}
    \label{fig: sim_opt_pd}
    \vspace{-0.1cm}
    \end{figure}

Since the PD controller’s feedback gains were deliberately set to low values in the previous simulation tasks, this may have led to an underestimation of the standalone PD controller’s capabilities based on the earlier simulation results. To address this potential misleading, we conducted an additional test using a better-selected feedback gain setting of $K_P = 1\times10^{-2}$ and $K_D = 1\times10^{-4}$. The results of using the three control methods in tracking a complexly varying signal with the optimized gain setting are shown in Fig.~\ref{fig: sim_opt_pd}. 

The results indicate that, with the optimized PD feedback gains, both the Linear+PD and the PD methods exhibited enhanced performance compared to the results in the previous tests. However, due to the linear model's limitations in controlling the nonlinear system, the Linear+PD method still exhibited significant control errors. Therefore, the standalone PD controller showed performance similar, and even slightly better, to that of the Linear+PD method in this test. Like the comparative methods, the ESN+PD also showed improved performance, with reduced control errors compared to former tests. Similar to the previous evaluations, the ESN+PD outperformed the other two methods, exhibiting the best performance with notably better constrained control errors.

\section{Experiments}

\subsection{Experimental Setup}

\begin{figure}[tb]
    \centering
\includegraphics[width=0.9\linewidth]{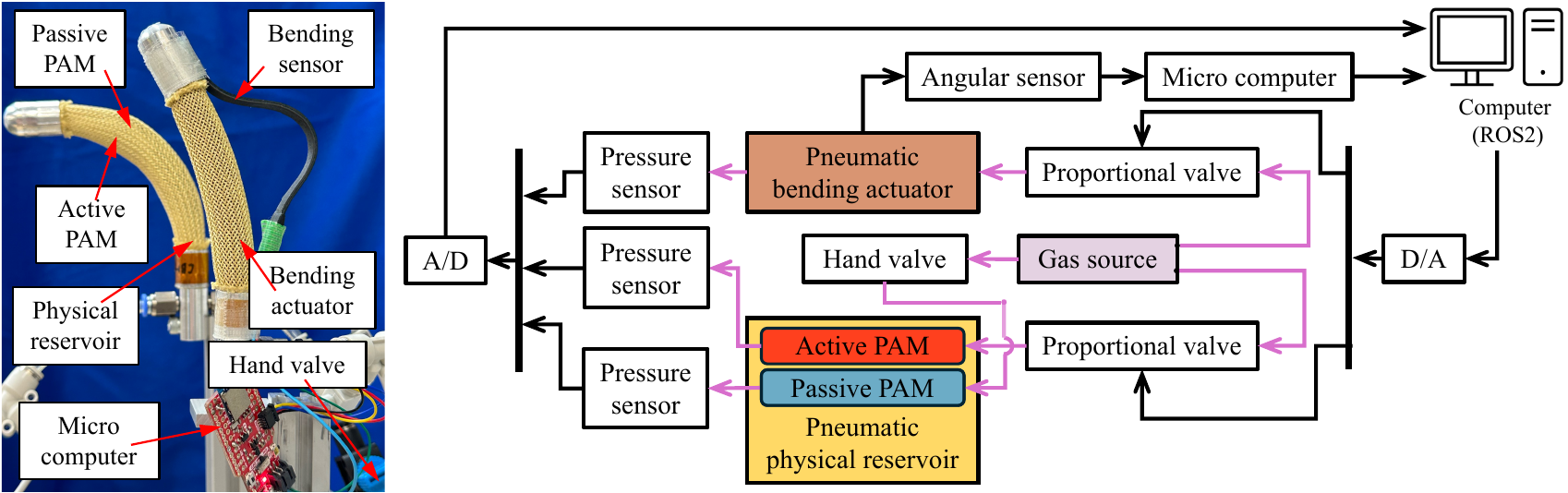}
    \caption{Experimental apparatus. Pink and black lines indicate the pneumatic and electronic circuits, respectively.}
    \label{fig: exp_app}
    \vspace{-0.2cm}
\end{figure}

We conducted experiments to validate the performance of the online learning control system, with the PRC model serving as the reservoir part in controlling the pneumatic soft actuator's bending output to track various reference signals. The control system follows that in Fig.~\ref{fig: system}, with the dual-PAM soft actuator in Fig.~\ref{fig: reservoir} as the physical reservoir. The experimental setup is shown in Fig.~\ref{fig: exp_app}, it consists of the pneumatic bending actuator and the dual-PAM soft actuator (both provided by Bridgestone), two proportional valves (Festo MPYE-5-M5-010B), an angular sensor (BENDLABS-1AXIS), three pressure sensors (SMC PSE540A-R06), and a ROS2 computer with a Contec AI-1616L-LPE analog input board and a Contec AO-1608L-LPE analog output board.

In the experiments, all signals were refreshed at 200 Hz, corresponding to a sampling interval $\tau$ of $5\times10^{-3}$ [s]. The hyperparameters were tuned through trial and error. The tap size $\delta$ was set to 50, and $n_u$ was set to 5. The RLS algorithm’s learning rate $\alpha$ was 1, and the forgetting factor $\lambda$ was $1-1\times10^{-6}$. The filter factor $\epsilon$ was 0.01, and the conversion factor $K_\text{in}$ was 7.0 [kPa/deg]. Like the simulations, to better demonstrate the feedforward PRC model’s effectiveness, the PD controller’s feedback gains were deliberately set to low values of $K^P=1\times10^{-2}$ [kPa/deg] and $K^D=1\times10^{-4}$ [kPa$\cdot$s/deg]. For the two PI controllers respectively depicted in Fig.~\ref{fig: esn_prc}(b) and Fig.~\ref{fig: system}, the P gains were set to $5\times10^{-2}$ [V/kPa], and the I gains were $5\times10^{-4}$ [V/kPa$\cdot$s]. The noise $\tilde{\cdot}$ inherently existed in the bending sensor, whose readings were used for feedback control and result illustrations. The saturation function $f_\text{sat}(\cdot)$ in experiments was defined as:
\begin{equation}
    f_\text{sat}(u_{k}) = \min\left(\max(u_{k}, \ 0), \ 400\right).
\label{eq: exp sat}
\end{equation}

Similar to our simulations, we also tested the performance of the Linear+PD and the PD control methods for performance comparison. The proposed control system, which combines the PRC model with the PD controller, is referred to as the PRC+PD method in the subsequent discussions.

\subsection{Experimental Results}

\begin{figure}[tb]
    \centering
\includegraphics[width=0.9\linewidth]{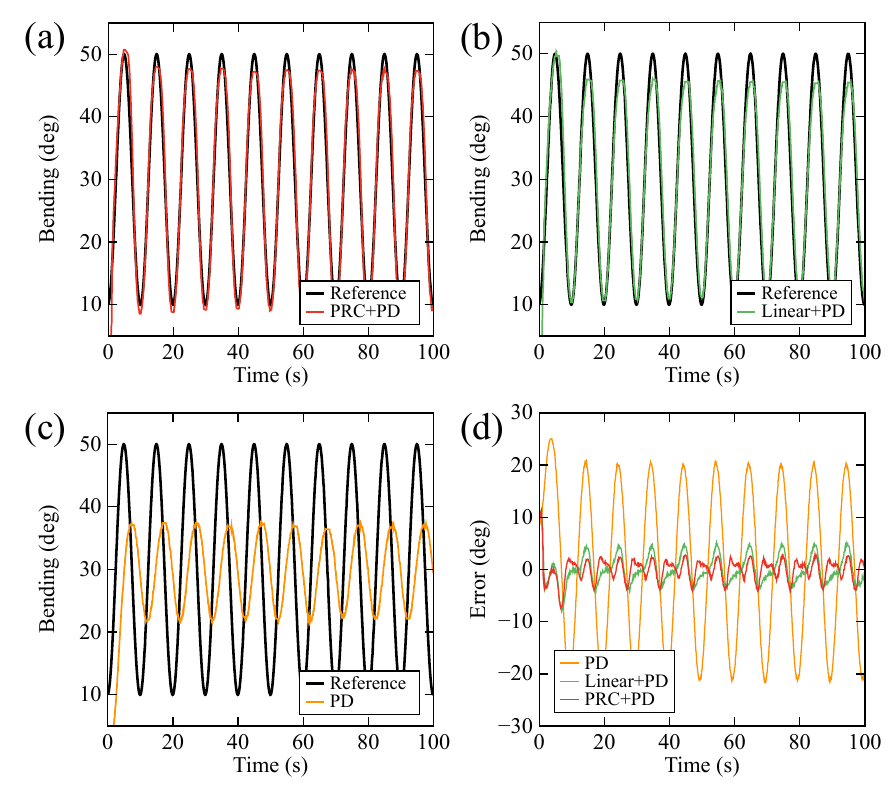}
    \caption{Experimental results of tracking a 0.1 Hz sine signal: (a) PRC+PD; (b) Linear+PD; (c) PD; (d) Control errors of different methods.}
    \label{fig: 0.1}
    \vspace{-0.2cm}
\end{figure}

\begin{figure}[tb]
    \centering
\includegraphics[width=0.9\linewidth]{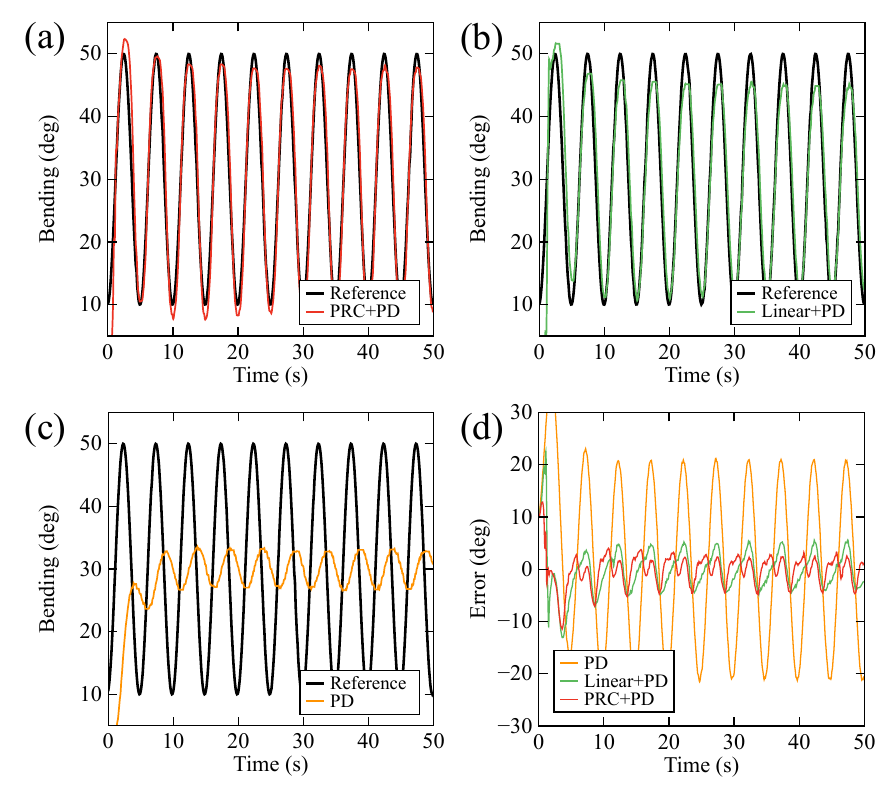}
    \caption{Experimental results of tracking a 0.2 Hz sine signal: (a) PRC+PD; (b) Linear+PD; (c) PD; (d) Control errors of different methods.}
    \label{fig: 0.2}
    \vspace{-0.2cm}
\end{figure}

We tested the performance of the three methods in controlling the pneumatic soft actuator's bending angle to track sine waveform reference signals ranging from 10° to 50° with frequencies of 0.1 Hz and 0.2 Hz. The obtained tracking outcomes are exhibited in Fig.~\ref{fig: 0.1}, and Fig.~\ref{fig: 0.2}, respectively.

The results indicate that, for both tested frequencies, the PD control method was unable to track the reference signal due to its slow response, a consequence of the low feedback gain settings. Although the Linear+PD and PRC+PD online learning control methods exhibited significant control errors at the initial stage—attributed to the undesired outputs of the completely untrained feedforward model at the beginning period—they both achieved rapid convergence from these significant initial errors and closely followed the reference signal. As shown by the green lines, the Linear+PD method failed to track the reference signal at the upper turning points. This failure can be attributed to the hysteresis dead zones present in the pneumatic soft bending actuator at the state transitions from pressurization to depressurization (corresponding to the upper turning points) \cite{shen2023trajectory}, which the linear feedforward component of the Linear+PD control system could not adequately compensate for. In contrast, while red lines indicate that the PRC+PD method also encountered such a challenge at these upper turning points, it demonstrated better tracking performance than the Linear+PD method, with a broader range of the reference signal's variations being covered. At the same time, due to this broader output motion range, the PRC+PD control system also illustrated some overshooting at the lower turning points.

\begin{figure}[tb]
    \centering
\includegraphics[width=0.9\linewidth]{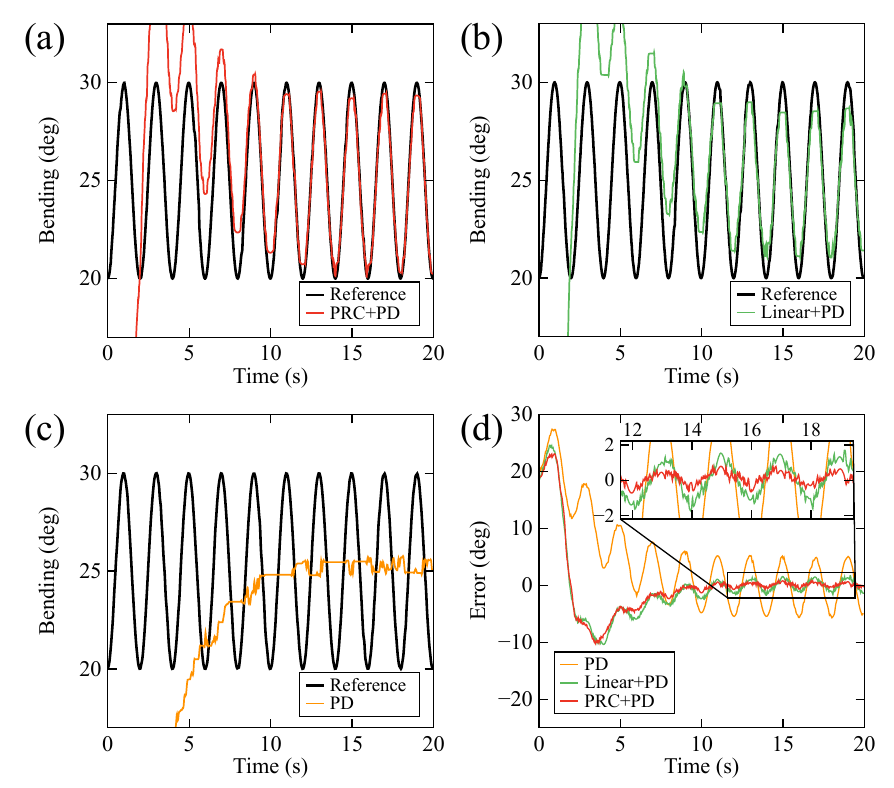}
    \caption{Experimental results of tracking a 0.5 Hz sine signal: (a) PRC+PD; (b) Linear+PD; (c) PD; (d) Control errors of different methods.}
    \label{fig: 0.5}
    \vspace{-0.2cm}
\end{figure}

Additionally, we evaluated the performance of different control methods in tracking a sinusoidal reference signal with a smaller motion range of 10° and a higher frequency of 0.5 Hz, the results are shown in Fig.~\ref{fig: 0.5}. Similar to the previous tests, the PD control method failed to follow the reference signal, and both the PRC+PD and Linear+PD methods exhibited substantial control errors during the initial period. The Linear+PD method also showed limitations in covering the full range of the reference signal during the steady-state tracking phase after the initial phase. In contrast, the PRC+PD method effectively controlled the soft actuator's bending output to closely follow the reference signal in the steady tracking period. The zoomed-in view of the control errors in Fig.~\ref{fig: 0.5}(d) further highlights that the PRC+PD method achieved a notably narrower error range compared to the Linear+PD method during the steady tracking period.

\begin{figure}[tb]
    \centering
\includegraphics[width=0.9\linewidth]{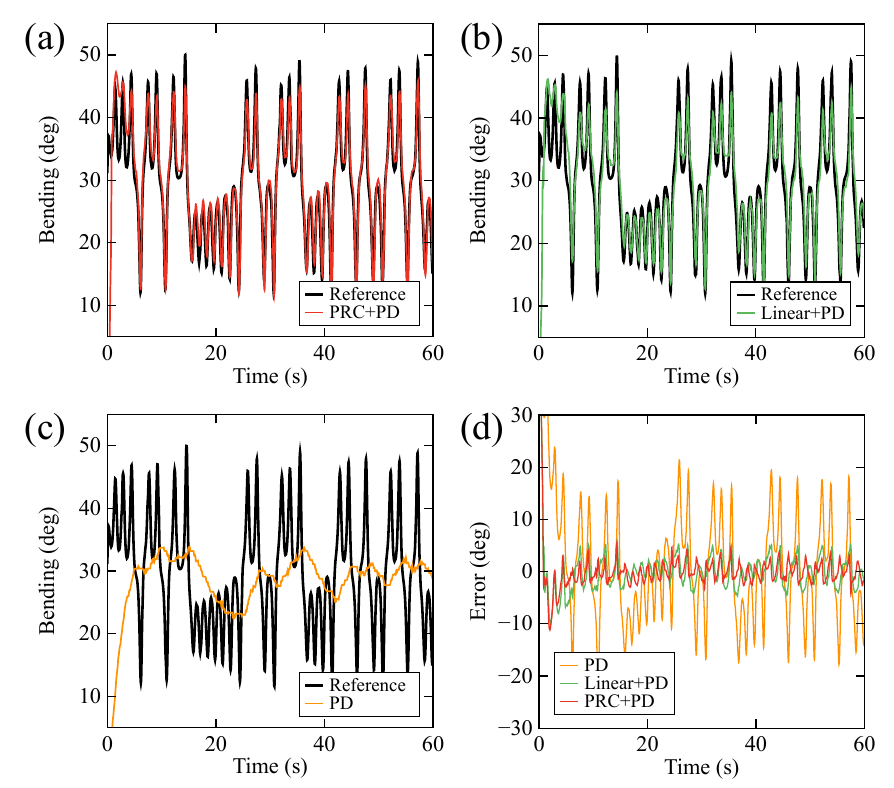}
    \caption{Experimental results of tracking a complex reference signal: (a) PRC+PD; (b) Linear+PD; (c) PD; (d) Control errors of different methods.}
    \label{fig: complex}
    \vspace{-0.3cm}
\end{figure}

Furthermore, we tested the performance of the three control methods in tracking a non-periodic complex reference signal, with the results shown in Fig.~\ref{fig: complex}. The experimental outcomes are consistent with those observed in previous tests tracking periodic reference signals. The PD control method was unable to perform adequately due to the intentionally low feedback gain settings. Both the Linear+PD and PRC+PD methods successfully tracked the reference signal after an initial phase characterized by significant control errors. The PRC+PD method demonstrated superior performance compared to the Linear+PD method, evidenced by its better capturing of the full range of the reference signal’s variations.

We numerically compared the performance of the three control methods using the root mean square error (RMSE). The RMSE [deg] for the tracking control result with $K$ total samples is calculated as $\sqrt{\sum_{k\in K}(y^d_k-\tilde{y}_k )^2/ K}$. Considering the significant tracking errors observed during the initial phase, only the data from the steady-state tracking phase after the 10-second initial period in each test were adopted. The results are shown in Table~\ref{tab: exp rmse}, with the best results for each test highlighted in bold. The results indicate that the proposed PRC+PD method achieved lower RMSE values than the Linear+PD method by an average reduction of over $37\%$ across all tested reference signals. The comparison results highlight the performance improvement brought by incorporating the pneumatic physical reservoir into the online learning control system to manage the nonlinear soft actuator's movements.

\begin{table}[t]
\centering
\caption{RMSE comparison of different control methods.}
\label{tab: exp rmse}
\renewcommand{\arraystretch}{0.8}
\begin{tabular}{lcccc}
\toprule
{Control method} & {Fig.~\ref{fig: 0.1}} & {Fig.~\ref{fig: 0.2}} & {Fig.~\ref{fig: 0.5}} & {Fig.~\ref{fig: complex}}\\
\midrule
PD      & 14.6  & 14.8 & 3.68 & 8.97    \\
Linear+PD   & 2.47 & 3.01 & 0.955 & 2.42   \\
PRC+PD   & \textbf{1.75} & \textbf{2.21} & \textbf{0.385} & \textbf{1.59}   \\
\bottomrule
\end{tabular}
\vspace{-0.4cm}
\end{table}

\section{Conclusion}

This paper presented an RC-based online learning control system design used to control the motion of a pneumatic soft actuator. The proposed control system integrated an RC model as a feedforward component with a PD controller for feedback adjustment. The RC model can be realized by either a conventional ESN or a PRC that utilizes physical systems for computation. The proposed control system's performance was evaluated by both simulations and experiments, where comparisons with a linear feedforward model highlighted the effectiveness of the RC-based feedforward model in controlling the output of nonlinear plants such as soft actuators. In simulations, an ESN model was used as the control system's RC component to regulate a simulated nonlinear system's output tracking various reference signals. The results demonstrated that the RC model achieved smaller control errors compared to the linear model. In the experiments, a PRC model based on a dual-PAM soft actuator was employed as the RC component to regulate the bending angles of another pneumatic soft actuator. The experimental results similarly confirmed the PRC model's superior performance, with an average of over $37\%$ reduced RMSE than the linear model.

The proposed PRC online learning control system features a more compact structure than existing designs and uses a physical system for computation. This compact architecture, with the integration of a physical system, offers a computationally lightweight solution for controlling soft actuators, making it advantageous for real-time deployment on edge devices with limited computational resources \cite{moon2019temporal}. It is noted that the simulation in this study employed a simple nonlinear plant as the target system. With advancements in soft robot simulation platforms, future work could explore applying the proposed control system for controlling simulated soft actuators, thus enabling a more comprehensive evaluation.

\FloatBarrier

\end{document}